\documentclass[a4paper,conference]{IEEEtran}

\ifCLASSINFOpdf
    \usepackage{subcaption}
\else
\fi
\usepackage{lipsum}
\usepackage{graphicx}

\graphicspath{ {./images/} }

\usepackage[usenames,dvipsnames]{color}
\definecolor{cb_red}{rgb}{0.89,0.1,0.11}
\definecolor{brown}{rgb}{0.59, 0.29, 0.0}
\definecolor{orange}{rgb}{1., 0.5, 0.25}

\begin{document}
%
\title{End-to-End Hierarchical Relation Extraction for Generic Form Understanding}


\author{\IEEEauthorblockN{Tuan Anh Nguyen Dang, Duc Thanh Hoang, Quang Bach Tran, Chih-Wei Pan, Thanh Dat Nguyen}
\IEEEauthorblockA{Cinnamon AI Labs\\
Hanoi, Vietnam\\
Corresponding author email: kristopher@cinnamon.is}}

\maketitle

\begin{abstract}

Form understanding is a challenging problem which aims to recognize semantic entities from the input document and their hierarchical relations.
Previous approaches face significant difficulty dealing with the complexity of the task, thus treat these objectives separately. To this end, we present a novel deep neural network to jointly perform both entity detection and link prediction in an end-to-end fashion.
Our model extends the Multi-stage Attentional U-Net architecture with the Part-Intensity Fields and Part-Association Fields for link prediction, enriching the spatial information flow with the additional supervision from entity linking.
We demonstrate the effectiveness of the model on the \textit{Form Understanding in Noisy Scanned Documents} \textit{(FUNSD)} dataset, where our method substantially outperforms the original model and state-of-the-art baselines in both Entity Labeling and Entity Linking task.
\footnote{https://github.com/hoangthanh283/MSAUPAF.git}

\end{abstract}


%
\IEEEpeerreviewmaketitle

\section{Introduction}
Administrative documents are one of the most widely used medium for data storage and communication, which raises the need for an automated solution for form understanding. However, this task remains largely difficult due to the layout variations between forms for different purposes, organizations, etc. Owning to its complexity, classical works like heuristic-based \cite{Oro2009}, \cite{iesurvey_simoes2005}, or recent deep learning-based methods \cite{Palm}, \cite{Liu2019} often break the problem into several sub-components: text-line detection, entity recognition, relation extraction, etc. 
The rationale of the division is that we may have better control of the input-output correspondence while cascading information for later steps. 
Following \cite{Jaume2019}, the problem of Form Understanding can be formulated as 3 main tasks: \textit{Word Grouping}, \textit{Entity Labeling} - detection of questions (keys) and answers (values), and \textit{Entity Linking} - their linking. 
Within this scope of this paper, we take another look at this prevalent problem by proposing an end-to-end model to solve both \textit{Entity Labeling} and \textit{Entity Linking} task at the same time. 

Aside from the aforementioned heuristic-based methods for document information extraction (i.e using hand-crafted rules to build up and extract tables and identifying key-value pairs), deep-learning-based methods are also received much attention during the recent years \cite{Johnson2018, Newell2017, Yang2018}.
Although deep-learning-based methods have achieved state-of-the-art accuracy recently in both entity segmentation in documents \cite{Madashi2019}, link prediction on large social network graph \cite{Zhang}, and matching between human pose joint \cite{Cao2016a}, there have been few works on end-to-end models performing both entity extraction and link prediction of a document at the same time. Our hypothesis is that while being advantageous for intermediate supervision, the division of architecture to different stages are not necessary for optimal speed and performance. As they heavily depend on each other as a flow, if the previous steps like entity recognition do not perform well, it would severely affect the performance of the later entity linking stage by error accumulations. 

In this paper, we present the extension of Semantic Entity Labeling with relational linking by adapting the Part-Association-Field mechanism for mutual relations extraction among entities.
Our work is based on the previously proposed Multi-Stage Attentional U-Net (MSAU) architecture \cite{Madashi2019} for its proven effectiveness in entity segmentation from the char-grid embedding \cite{Katti2018}. We go beyond its limitation of having no direct supervision from the linking between entities by adding Part-Intensity Fields and Part-Association Fields (PIF-PAF) \cite{Kreiss2019} which results in an additional two-head branch for entity linking and prediction from document images. 
In addition to original segmentation output, the Part-Intensity Field head produces of confidence score of each entity's key points, while the Part-Association Fields head allows the associations between entities. 
Apart from the major addition of PIF-PAF module, our work also improved upon MSAU by adding Corner Pooling \cite{Law2018} to solve the issue of long-range distance between entities, enabling the wider-range propagation of information both horizontally and vertically. 
The Coordinate Convolution (CoordConv) \cite{Liu2018a} is also included to incorporate translational signals to the model.
The effectiveness of these modules will be shown in Section \ref{sec:experiments}.

We demonstrate the capability of our proposed model on the \textit{Form Understanding in Noisy Scanned Documents} dataset (\textit{FUNDS}) \cite{Jaume2019}, focusing on the \textit{Entity Labeling} and \textit{Entity Linking} task. While the line of work for Entity Labeling is actively developed in \cite{Liu2019, Madashi2019, Palm, Palm2018}, for the combined \textit{Entity Labeling} and \textit{Entity Linking} task, the methods reported in \cite{Jaume2019}, \cite{Xu2019}, \cite{Pramanik2020} only achieved limited performance due to the bottleneck in exploiting both semantic and spatial features and the highly imbalanced links between entities. Under those challenging conditions, experiments demonstrate that our approach outperforms various strong baselines (including Multi-layered Perceptron (MLP) with BERT, recently proposed MSAU \cite{Madashi2019}, LayoutLM \cite{Xu2019}, and Graph-based methods \cite{Such2017}), achieving state-of-the-art performance. 

In short, we summarize our contribution list:
\begin{itemize}
    \item We proposed an end-to-end architecture combining semantic entity extraction and link prediction for form understanding, which incorporates the Part-Association Field and Part-Intensity Field to the baseline MSAU model.
    \item We improved upon the original architecture of MSAU by adding Corner Pooling and Coordinate Convolution to aid spatial context modeling, which has been reaffirmed by our empirical experiments.
    \item Ablation studies show that utilizing the relations supervision can improve segmentation results for entity extractions in comparison with other baselines.
\end{itemize}

The rest of the paper is organized as follows: Section \ref{sec:related_works} describes the related works, followed by that is the proposed method in Section \ref{sec:Methodology}, the experiments in Section \ref{sec:experiments}.

\section{Related Work}
\label{sec:related_works}
Our works focus on incorporating both information extraction from document images and Entity Linking into one model. The former consists of several sub-themes: heuristic methods, semantic segmentation on images, graph-based information extraction. 
The Entity Linking part of our model in particular also incorporates techniques from human pose estimation and keypoints association, which we will describe briefly below and in detail in Section \ref{sec:Methodology}.

\textbf{Heuristic methods:} Heuristic methods for structured data extraction are either concerned with the layout analysis problem before the information extraction, the first stage (Word Grouping) or the third stage (Entity Linking). 
Prior to the information extraction stage, \cite{Bukhari2011, Ahn2017} tries to build up the components using carefully designed feature engineering (ridges and whitespaces \cite{Bukhari2011}, connected component analysis \cite{Ahn2017}) for text lines segmentation from images.
This step is important because its performance determines the accuracy for the rest of the pipeline, less-precise segmentation requires more robustness (for over/under segmentation errors) in later steps. 
Heuristic methods for table segmentation and extraction \cite{Oro2009, Chakrabarti2008} can be used in either the first or the third step in form understanding task.
While \cite{Oro2009} separated table detection tasks into specific stages: Clustering words to text lines, merging text lines into tables by heuristic rules (Same task as Word-Grouping \cite{Jaume2019}). \cite{Chakrabarti2008} represents text lines as graph nodes and finds the linking between them by calculating the edged-weights (the cost of two nodes belonging to the same types).
Since these algorithms are designed tailoring a certain condition (i.e specific type of forms, layouts, required information), their scalability/generalizability towards large datasets is limited. 
The MSAU \cite{Madashi2019} and our model, however, are based on character-level mask, which leads to naturally more robust to these kinds of errors \cite{Madashi2019}, the deep segmentation model also improves generalizability.

\textbf{Deep learning-based segmentation and Link prediction:}
Early work in information extraction from document images formalize the problem as a segmentation problem based on either Fully Convolutional Networks \cite{Long2015} or U-Net \cite{Ronneberger2015, Oliveira2018} applied fully convolutional network to directly segment regions of interest from historical documents. 
\cite{Palm} proposed architecture takes the spatial structure into account by using convolutional operations on the concatenated document text and image modalities, the text modality is represented by embedding the extracted text in a spatial grid.
Our work built upon \cite{Madashi2019} is more aligned with recent end-to-end systems incorporating both semantic structure and spatial image features \cite{Katti2018, Palm2018}. 
Similar to \cite{Madashi2019}, char-grid \cite{Katti2018} is also employed instead of sub-word, word, or sentence level embedding.
The character-level embedding has two advantages: embedding character-level is less memory-consuming and has been proven to be robust against potential accumulated text-line OCR and segmentation errors \cite{Madashi2019}. 

\textbf{Information extraction with Graph Neural Networks:}
Recently, there are several works on modeling the entity's relationship based on graph convolution networks \cite{Liu2019, Lohani2019} which have achieved promising results. 
\cite{Liu2019} introduced edge embeddings into the graph convolution network, which models the relationship between vertices directly while both works \cite{Lohani2019} used the nearest spatial distanced entity as a key to determining the linking for graph building. Graph Neural Networks in general, have also been used for link-prediction problems (which is close to the Entity Linking stage) \cite{Zhang}. 
However, to the best of our knowledge, there have not been any attempts to incorporate both the tasks of information extraction (formalized as node  classification in \cite{Liu2019, Lohani2019}), and Entity Linking.
We believe that this line of direction is very potential due to the capability of incorporating both semantic, spatial information. However, the implementation of these methods, have yet to reach the best possible performance under the condition of erroneous output from earlier steps (text lines recognition and optical character recognition).
The char-grid \cite{Katti2018} is naturally robust to these types of errors as stated in \cite{Madashi2019}, this is the reason we chose to develop our work of Entity Linking upon the line of segmentation over char-grid.

\textbf{Human pose estimation:} As mentioned earlier, our improvement for the incorporation of Entity Linking is highly relevant to the technique used in Human Pose Estimation, applied in a sense that linking between entities can be similar to linking between joints of human poses.
In the context of human pose estimation, there are mainly two approaches: top-down and bottom-up. The top-down approach starts by identifying and localizing individual person instances, followed by single pose estimation \cite{Papandreou2017, Xiao2018}.
Although this approach makes the human pose estimation is more straightforward, it might not be applicable to our case because unlike human joints which are often close to each other, key-value pairs can be distanced apart. 
Furthermore, a single key in our case can match multiple values, making the process of pinpointing accurately bounding box of all pairs complicated due to overlapping. Hence we found this is not for capturing the entity's relation in the documents. 

In the bottom-up approach, some recent works use greedy decoders in combination with additional tools to generate person instance as in \cite{Newell2017} propose associative embedding to identify key points from the same person, \cite{Cao2016a} use Part Affinity Fields to infer human poses. Inspired from \cite{Cao2016a} the PIF-PAF \cite{Kreiss2019} incorporates human body parts with a Part-Association field, which yields better quality results, and drastically reduces prediction time. The characteristic of this method involving bipartite matching, we believe, is similar to the match between key-value pairs. Inspired from this line of work, we incorporate Part-Association Fields as part of the model to encode of joint association all entities in the document. 

In short, we built our work upon the line of deep-learning-based models, in particular, MSAU \cite{Madashi2019}. Considering the advantageous characteristics of Part-Association Fields in matching human joints, we incorporated the technique into our model for end-to-end key-value detection and matching.
On top of that, as we consider key-point is a text line's corner to localize and the text-line size of text-lines can vary greatly or be occluded, also does the translational information (i.e Title is likely on top of the page). Of which, Corner Pooling from the work of Corner Net \cite{Law2018} has been proven to be useful in dealing with text line size variations. For the incorporation of translational information, which is potentially useful for link-prediction, Coordinate Convolution \cite{Liu2018a} mechanism has also been proven to be effective. We incorporate both of these \cite{Liu2018a, Law2018}, in our model as described in Section \ref{sec:Methodology}.

\section{Methodology}
\label{sec:Methodology}
\begin{figure*}[ht]
    \centering
    \includegraphics[width=400pt, height=250pt]{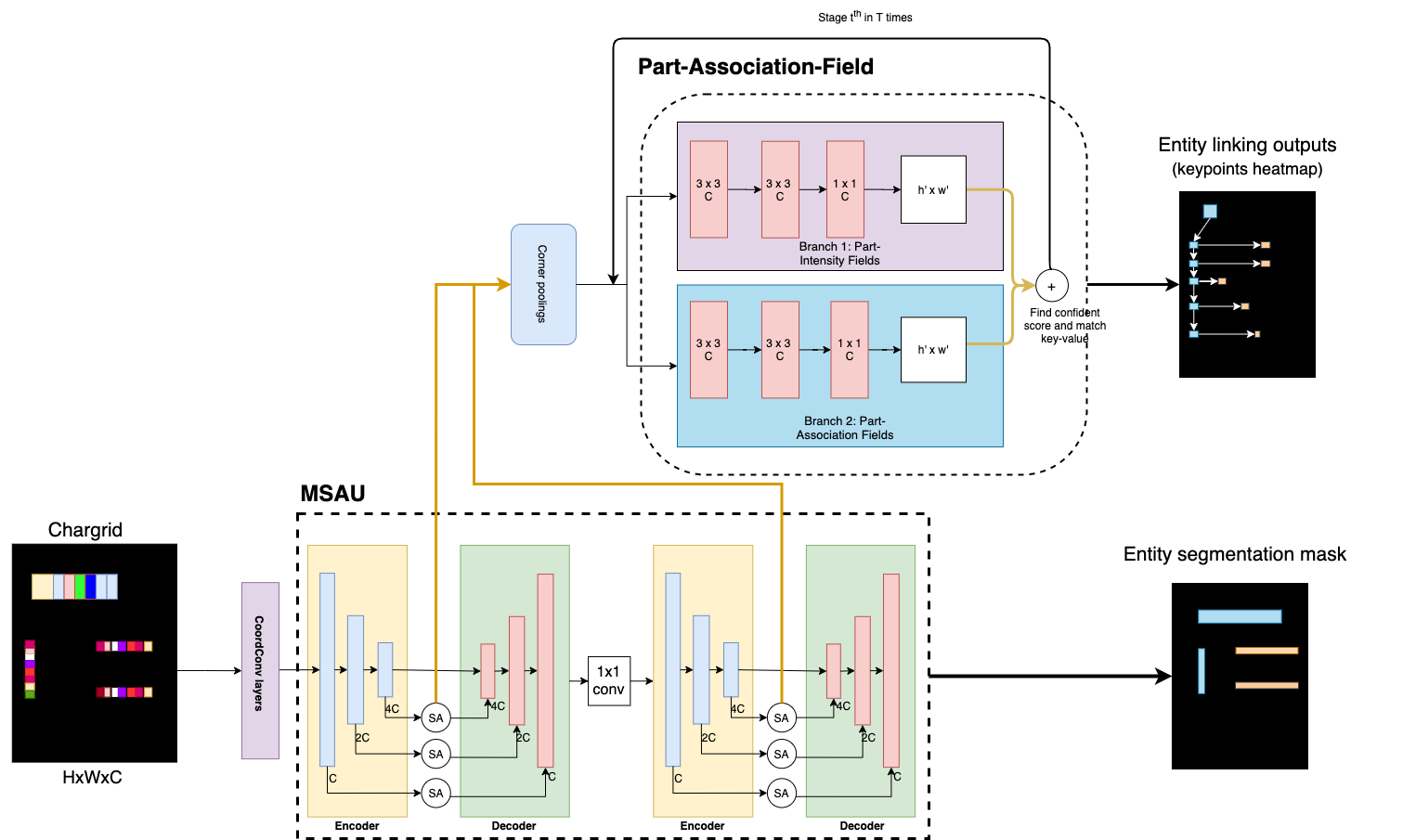}
    \caption{\small{The architecture of MSAU-PAF prediction. The MSAU Block is based on MSAU \cite{Madashi2019} with adding some components which are Attention layers \cite{Wang-non-local} and CoordConv layers \cite{Liu2018a} to exploit spatial information from the image. After this stage, the prediction is forwarded through a Corner Pooling and taken as input of Part-Association-Field. The Part-Association-Field block is a two-heads net with Part-Intensity Fields (PIF) and Part-Association Fields (PAF) and produce the linking prediction in document image.}}
    \label{fig:architecture}
\end{figure*}
An overview of the proposed method, MSAU-PAF, is depicted in Figure \ref{fig:architecture}, it takes the same input as MSAU, namely the char-grid (see Subsection \ref{subsec:char-grid}). The model aims to handles Entity Recognition and Entity Linking in an end-to-end fashion. 
The Entity Recognition output is the same as MSAU (Subsection \ref{subsec:msau}), for which the baseline architecture of MSAU is incorporated and nearly unchanged except the addition of CoordConv (Subsection \ref{subsec:coordconv}. 

    The model is also expected to produce the linking between them correspondingly with class and relation of entities in the dataset. For this task Part-Intensity Fields and Part-Association Fields (PIF-PAF) modules \cite{Kreiss2019} are employed, MSAU-PAF interfaces between MSAU and PIF-PAF by concatenating output from bottleneck layer of MSAU and make use of Corner Pooling (see Subsection \ref{subsec:pifpaf}).

The head of MSAU will be used to segment entity classes while the other two head networks (PIF-PAF) are used as follows: PIF head is used for predicting confidence, precise location, and size of keypoints, while PAF head predicts associations between keypoints (entities), called Part-Association Field (PAF). 

\subsection{Char-grid}
\label{subsec:char-grid}
We adopt the char-grid representation \cite{Katti2018} as the method of embedding each character's information into a feature map region that the character occupies. 
This embedding method allows the incorporation of both character position and semantic information at the same time. 
Thus it is proven to be effective in information extraction from document images \cite{Katti2018, Madashi2019}.
We obtain the char-grid $CM \in \{0, 1, ..., N_{char}\}^{H \times W}$ similar to \cite{Madashi2019}, where $N_{char}$ is the number of chosen characters, $H$ and $W$ are the height and width of the input, respectively. 

\subsection{Multi-stage Attentional U-Net}
\label{subsec:msau}

We choose MSAU \cite{Madashi2019} as baseline due to several advantages that give us even more supervision over the model output and the evaluated model's empirical performance.
MSAU \cite{Madashi2019} improved upon Coupled U-Net \cite{Tang2019-cU-Net} with several techniques such as Non-local block \cite{Wang}, and Box Convolution \cite{box_conv_Burkov2018}.
The idea is to segment entities from char-grid with multiple intermediate outputs (segment key mask at one block and value mask at another) for better supervision.
As a result, MSAU has been proven to be effective in entity segmentation tasks \cite{Madashi2019}.
We extend the work of MSAU to integrate Entity Linking explicitly to the model, especially long-ranged semantic relations.
our improvement over the base line architecture is shown in MSAU block of Figure \ref{fig:architecture},
where we added the coordinate convolution layers \cite{Liu2018a} prior to the first U-Net block. 
The first U-Net block of MSAU is used for output key masks, while the second output full key-value segmentation.
The output of attention layers from the bottleneck (output from the yellow block in the diagram) is used to feed into corner-pooling modules, followed by PIF-PAF \cite{Kreiss2019} module to output the key-points confidence score and Entity Linking maps.
The rest of the architecture is left untouched.

\subsection{Part-Intensity Fields and Part-Association Field}
\label{subsec:pifpaf}
Inspire from recent human pose estimation advancement, the Part-Intensity Fields (PIF), and Part-Association Fields (PAF) \cite{Kreiss2019} are all confidence maps with vector fields. 
Each vector field composes a scalar component for confidence, size of the keypoint and vector components. 
The outputs from bottlenecks of each U-Net block of MSAU are concatenated and go through corner pooling before being fed to PIF-PAF.
Thus, the input of PIF-PAF block is a set of feature maps {F}. 
PIF-PAF has multiple stages, each stage is a PIF-PAF block, similar to each U-Net block of MSAU. 
The PIF will find and localize key points via predicting a composite field, containing confidence score $c$, a vector $(x, y)$ with spread $b$ and a scale $\sigma$, in each location $(i, j)$ in the output maps which can be written as $p^{ij} = \{p^{ij}_c, p^{ij}_x, p^{ij}_y, p^{ij}_b, p^{ij}_\sigma\}$. 
On the other side, the role of PAF would be to find two entities, which not necessarily stay close to each other and could be far away, and determine the confidence of their association. 
At every output location, PAFs will be represented as $a^{ij} = \{a^{ij}_c, a^{ij}_{x1}, a^{ij}_{y1}, a^{ij}_{b1}, a^{ij}_{x2}, a^{ij}_{y2}, a^{ij}_{b2}\}$, this help the PAFs have the capacity to estimate multiple entities' associations. By enjoying the fine-grained information from PIFs, the PAFs could utilize this to predict entity's locations and their association which have been shown to be efficient and robust in dense and occluded documents via our experiments.
\subsection{Coordinate Convolution}
\label{subsec:coordconv}
We take into account the fact that, since MSAU-PAF's inputs (character masks) are built from the output potentially erroneous text line recognition steps from scanned or camera-based document images, there might be variations in terms of text-lines scale, along with other distortions such as slight rotation, skewing, and non-aligned text-lines.

In order to improve MSAU-PAF's translational awareness in the entity's localization, Coordinate Convolution (CoordConv) \cite{Liu2018a} is applied. 
CoordConv adds two extra channels (in case of images) filled with coordinate information, concatenated at the end existing input feature map, allowing the model to learn varying degrees of translation-based cues. Zero weights for these channels would imply full translational invariance like standard convolution while any other value would lead the network to learn translation variance up to a degree which becomes a training parameter depending on the task. It is applied in both MSAU and the head networks.
\subsection{Corner Pooling}
Since text-lines and key-value pairs' distances vary in sizes, representing entities' key points as single pixels in feature maps might make it hard for the PAF heads to propagate information spatially between entities. 
In detail, intuitively, the receptive fields in normal convolution layers may not be able to effectively capture the local visual evidence in a long range distance. 
Furthermore, there exist cases where linked entities are far from each other, such as item values with respect to headers in a table. 
To address this issue, we apply Corner Pooling \cite{Law2018} to better gather all visual features in a long-distance by max-pooling horizontally towards the right and vertically towards the bottom of the feature maps.
The corner pooling encodes explicit prior knowledge about the objects' corners. 
Given $H \times W$ feature maps, the corner pooling layer first perform max-pooling on all feature vectors between $(i, j)$ and $(i, H)$ in $f_t$, resulting in a feature vector $t_{ij}$. Followed by that, all feature vectors between $(i, j)$ and $(W, j)$ in $f_t$ are max-pooled to a feature vector $l_{ij}$. Finally, $t_{ij}$ and $l_{ij}$ are added together.

\subsection{Training details}
\label{subsec:training_detail}
In this section, we describe the detail of the training stage, including data augmentation, model hyper-parameters, and the loss function. Our model is set up as follows:
\begin{itemize}
    \item \textbf{Data augmentation} We perform several data augmentation methods in order to increase the capacity of our model against the variety of layouts as follows: random character replacement to mimic OCR errors, random text box shifting, random affine transformation, random background padding. In addition, the text-regions in the input image are down-scaled so that the median text height is 3 pixels in the char-grid. 
    \item \textbf{Implementation detail} With char-grid input, we use the most frequency characters N = 90. We follow the MSAU's implementation with the number of U-Net blocks $n_{blocks} = 2$, the depth of the ResBlock $res\_depth = 2$. the number of downsampling blocks in an encoder $n_{downsampling} = 4$. Moreover, we also add CoordConv operation \cite{Liu2018a} to each layer of U-Net's encoder and use corner-net at the end of the final U-Net's decoder. As in Figure \ref{fig:architecture}, at the end of MSAU, the model is separated into two branches which are PAFs and PIFs.

    \item  \textbf{Loss function} The multi-task loss used in the training process is the sum of three smaller losses. We apply Cross-Entropy loss \cite{Madashi2019} at the output of MSAU block to guide the network to learn the segmentation mask between ground-truth label and predicted label. Moreover, to predict the confidence maps of entities in the first branch PIF and the PAFs in the second branch, we used two Laplace losses \cite{Kreiss2019} at each branch between the estimated predictions and the ground-truth maps. The Laplace losses, which is a modified $L_{1}$ type loss with injecting a predicted spread $b$ dependence, helps the network handle the diversity of scales of text-lines in the documents. The loss is formulated as below:
    \begin{equation}\label{my_fourth_eqn}
        L_{total} = \lambda_{1} L_{CE} + \lambda_{2} L_{Laplace-PIF} + \lambda_{3} L_{Laplace-PAF}
    \end{equation} with $\lambda_{1}$, $\lambda_{2}$ and $\lambda_{3}$, loss weights, are set to $1, 1e-2, 1e-2$ respectively.
\end{itemize}
\section{Experiments}
\label{sec:experiments}
\subsection{Dataset}

For evaluation, the FUNSD (Form Understanding in Noisy Scanned Document) dataset is used \cite{Jaume2019}.
The dataset consists of 3 different tasks: Text Detection, Text Recognition (Optical Character Recognition), and Form Understanding. 
Following the description in \cite{Jaume2019}, Form Understanding is further broken down into three sub tasks: Word Grouping, Semantic Entity Labeling and Entity Linking.

\textbf{Word Grouping} Word grouping concerns the clustering of words belong to the same entity. 
The task takes input as word locations, each as a tuple $(x1, y1, x2, y2)$ denoting the upper left corner's and lower right corner's image coordinates of each word's bounding box. 
It is required from the task to clusters the words that belongs to the same entity, namely, the system solving the challenge is expected to output the bounding box of each entity and the merged text from accumulated words.

\begin{figure}[ht]
    \centering
    \caption{The class distribution of entities in FUNSD dataset \cite{Jaume2019}. There are four main classes which are Header, Question, Answer and Other.}
    \includegraphics[width=\linewidth]{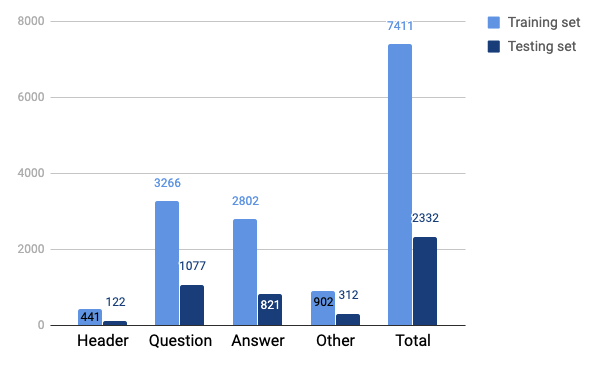}
    \label{fig:class_distribution}
\end{figure}

\textbf{Semantic Entity Labeling} The input of this task is the coordinate of text line (received from the output of word grouping tasks). This task requires classifying entities into one of four pre-defined categories: question, answer, header and, other. 
See Table~\ref{fig:class_distribution} for the distribution of classes.

\textbf{Entity Linking} The task takes accumulated output information from 2 previous tasks as input: coordinates of text-line regions, text-recognition result, and Entity Labeling. With each text line given an $ID$, the system is expected to output entities relations under the form of pairs $(id_1, id_2)$ where $id_1$ is the unique ID of entity labeled as question and $id_2$ is the ID of entity labeled as the answer.

The dataset contains 199 fully annotated forms with a variety of structure and appearance (e.g marketing, science reports, invoice), separated into two subsets which are training set and test set.
The statistics of the dataset can be seen in Table~\ref{table:datastatistic} and Figure \ref{fig:class_distribution}, further detail can be found in the description paper \cite{Jaume2019}.

\begin{table}
\centering
\caption{\label{table:datastatistic} The statistics of FUNSD dataset \cite{Jaume2019}: The table shows number of words and number of entities in total. And between entities, the relation between them are labeled by linking the "id" of entities}
\begin{tabular}{ccccc}
\hline
Split    & Forms & Words   & Entities & Relations \\ \hline
Training & 149   & 22, 512 & 7, 411   & 4, 236    \\ \hline
Testing  & 50    & 8, 973  & 2, 332   & 1, 076    \\ \hline
\end{tabular}
\end{table}%

\subsection{Experiment setup}
In this paper, we only focus on the tasks that belong to Form understanding. 
Due to the characteristic of the model not requiring entities to be grouped (first task output), the evaluation is done on the two sub-tasks: Semantic Entity Labeling and Entity Linking

For Semantic Entity Labeling, F1-score is employed for evaluation. The F1-score is calculated from the semantic segmentation output of the model using the box F1-Score introduced in \cite{Madashi2019} - a predicted box is marked as a correct detection if its $IoU$ ratio with one of the ground-truth boxes is larger than a certain threshold $0.8$.

The Entity Linking task remains the same. Given the output relations between by pair $(id_1, id_2)$, the system is benchmarked using the F1-score of all possible pairs as in \cite{Jaume2019}

For the task of Semantic Entity Labeling, these baselines are used:
\begin{itemize}
    \item \textit{MLP} - Multi-layered Perceptron (results are referenced directly from \cite{Jaume2019})
    \item \textit{MSAU} - Multi-stage Attentional U-Net \cite{Madashi2019}
    \item \textit{LayoutLM} - Pre-training of Text and Layout for Document Image Understanding \cite{Xu2019}
\end{itemize}

For Entity Linking, these baselines are used:
\begin{itemize}
    \item \textit{MLP} - Multi-layered Perceptron (results are referenced from \cite{Jaume2019}.
    \item \textit{GraphCNN with Link prediction.}
    \item \textit{Heuristics method.}
\end{itemize}

The baseline methods are tested with the normal input of each task (i.e MSAU only evaluated on one task of Semantic Entity  Labeling, GraphCNN with Link Prediction is run given assumed perfect output from Semantic Entity Labeling). Detail of the baseline methods can be referred to in the next section.
For MSAU-PAF, it is emphasized that no intermediate input is given between the first input and Entity Linking output.
We set up our baseline models (MSAU and Graph) and MSAU-PAF, the training processes are performed on a server with Tesla V100  (with 16GB memory GPU). The model is trained 200 epochs with a mini-batch size of 4. In order to optimize the model, we use the RMSProp method. The learning rate is set to 0.001 at the beginning and decreased every 10 epochs with a polynomial decay of 0.9. The result is shown in Subsection \ref{subsec:results}

\subsection{Baseline  methods}
\textbf{MLP} For Entity Labeling, the input of the model is features extracted from pre-trained model BERT \cite{Devlin2018} as a 733-dimensional vector and forward through two 500-units hidden layers with ReLU activation and the final output is put through a softmax layer. For the Entity Linking task, the input of the model is the concatenation of every pair of Semantic Entity.  

\textbf{MSAU} Following the settings of MSAU in \cite{Madashi2019}, 2 U-Net blocks are used, with $n_{downsampling}$ = 5 (5 down-sampling layers each block) and ${res\_depth}$ = 2 (2 residual convolutions in each down/up-sampling layer) and without Box Convolution. The MSAU-PAF shares the same structure with MSAU except that the output is modified as a binary classification.

\textbf{GraphCNN with Link prediction} the Graph Convolution Operations employed in the experiment is Spatial Graph Filtering, introduced in \cite{Such2017}, text lines are encoded from BERT with $733$-dimensional vector, combined with normalized coordinate $(x1, y1, x2, y2)$ to produce $737$-dimensional vector as input. 
The spatial graph of text lines is built from nearest aligned texts as an indication of having a spatial edge. 
Given $N \times F$ ($F = 737$) and $A \in \{0, 1\}^{N \times N \times L}$ ($L = 4$: Top, bottom, left, right) as input node features and adjacency matrix. The baseline model has 4 layers, encoding the input to $256, 128, 128$ sequentially, the final layer's outputs of the model for each node pairs is then fed to a single hidden layer with 500 units and output whether a link exists between two nodes.

\textbf{Heuristics method} The input of the model is the ground-truth class and coordinate of text-line regions. After that, several Rules-based are set up as follows: with an entity labeled as the answer, the model finds the nearest entity labeled as a question by measuring the Euclid distance between this entity and others ”question” entities and the result are pairs $(id_1, id_2)$ of entities with $id_1$ is the $ID$ of question text-line region and $id_2$ is the $ID$ of answer text-line region.

\subsection{Results}
\label{subsec:results}
\begin{table*}[ht] 
\caption{Result of MSAU-PAF: The first and second rows are the input and char-grid after being processed respectively. The third row shows the entity locations through bottom-left points and the fourth row visualizes the results of the model. The header, question, and answer are segmented by yellow, green, and blue color correspondingly. The links between entities (if existed) are visualized through orange lines.} \label{table:resultofmodel}
\small\centering
\begin{tabular}{cccccc}
Images
&\begin{tabular}{c}\includegraphics[width=2cm, height=3cm]{images/original_images/82253058_3059.png} \end{tabular}
&\begin{tabular}{c}\includegraphics[width=2cm, height=3cm]{images/original_images/82573104.png}
\end{tabular}
&\begin{tabular}{c}\includegraphics[width=2cm, height=3cm]{images/original_images/83624198.png}
\end{tabular}
&\begin{tabular}{c}\includegraphics[width=2cm, height=3cm]{images/original_images/83996357.png}
\end{tabular}
&\begin{tabular}{c}\includegraphics[width=2cm, height=3cm]{images/original_images/87137840.png}
\end{tabular}\\

Char-grid        
&\begin{tabular}{c}\includegraphics[width=2cm, height=3cm]{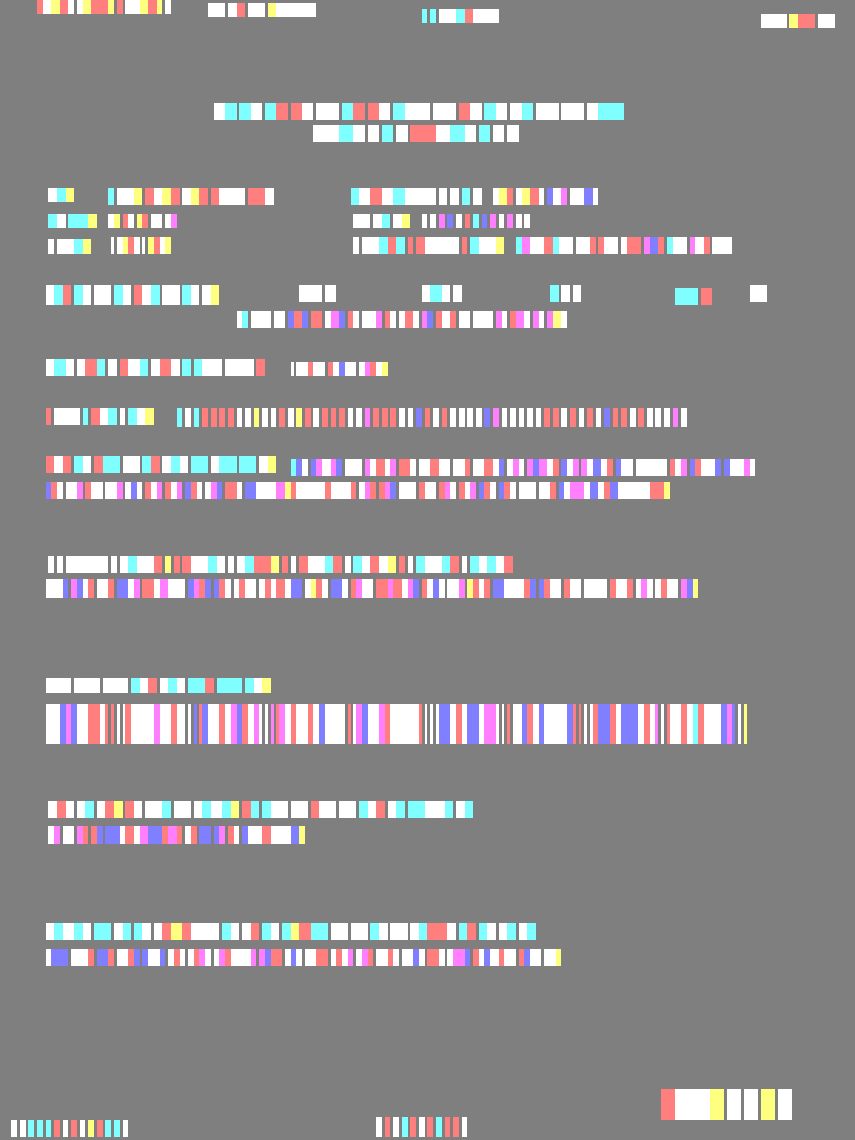}
\end{tabular}
&\begin{tabular}{c}\includegraphics[width=2cm, height=3cm]{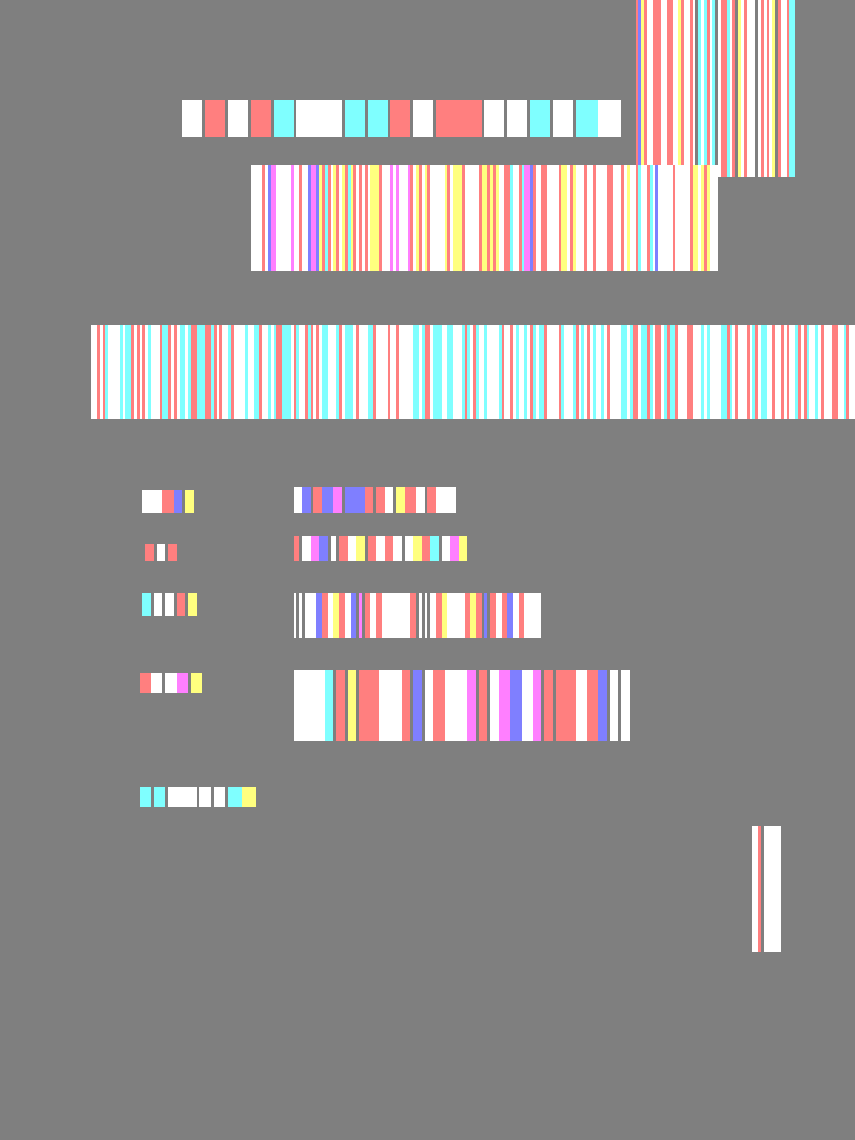}
\end{tabular}
&\begin{tabular}{c}\includegraphics[width=2cm, height=3cm]{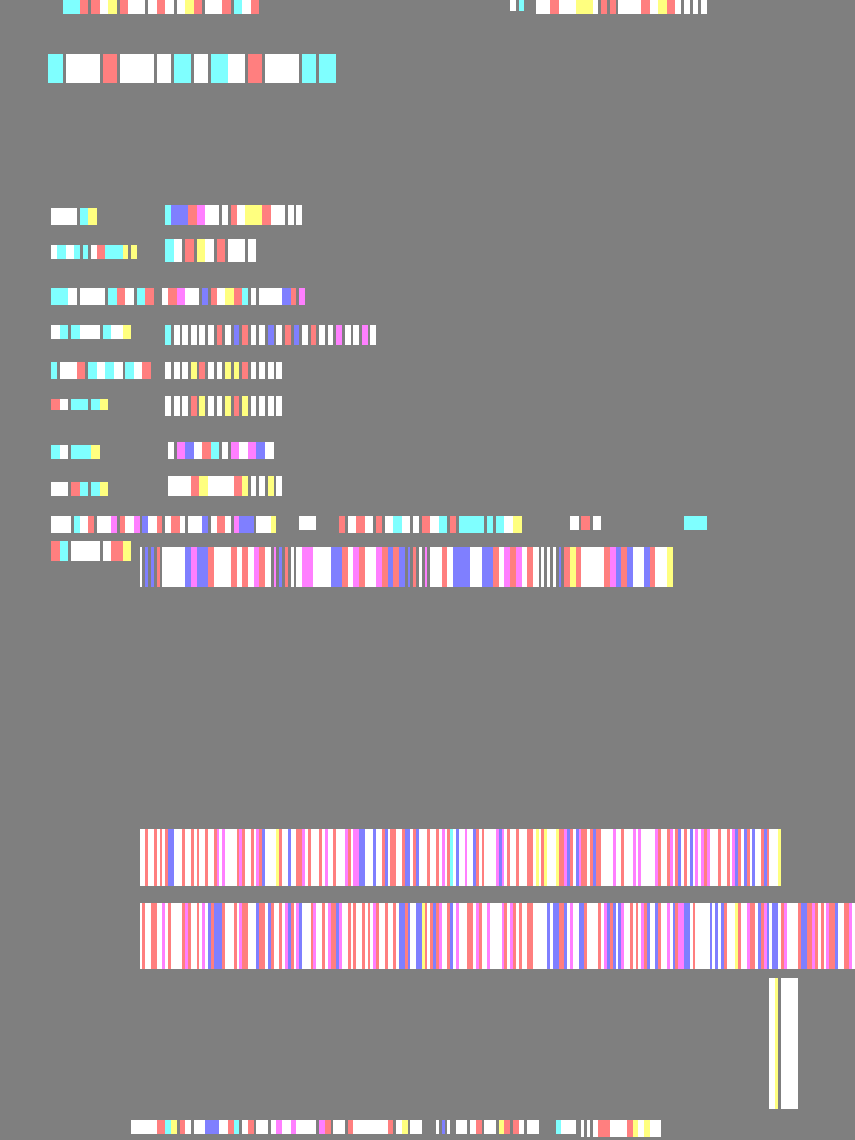}
\end{tabular}
&\begin{tabular}{c}\includegraphics[width=2cm, height=3cm]{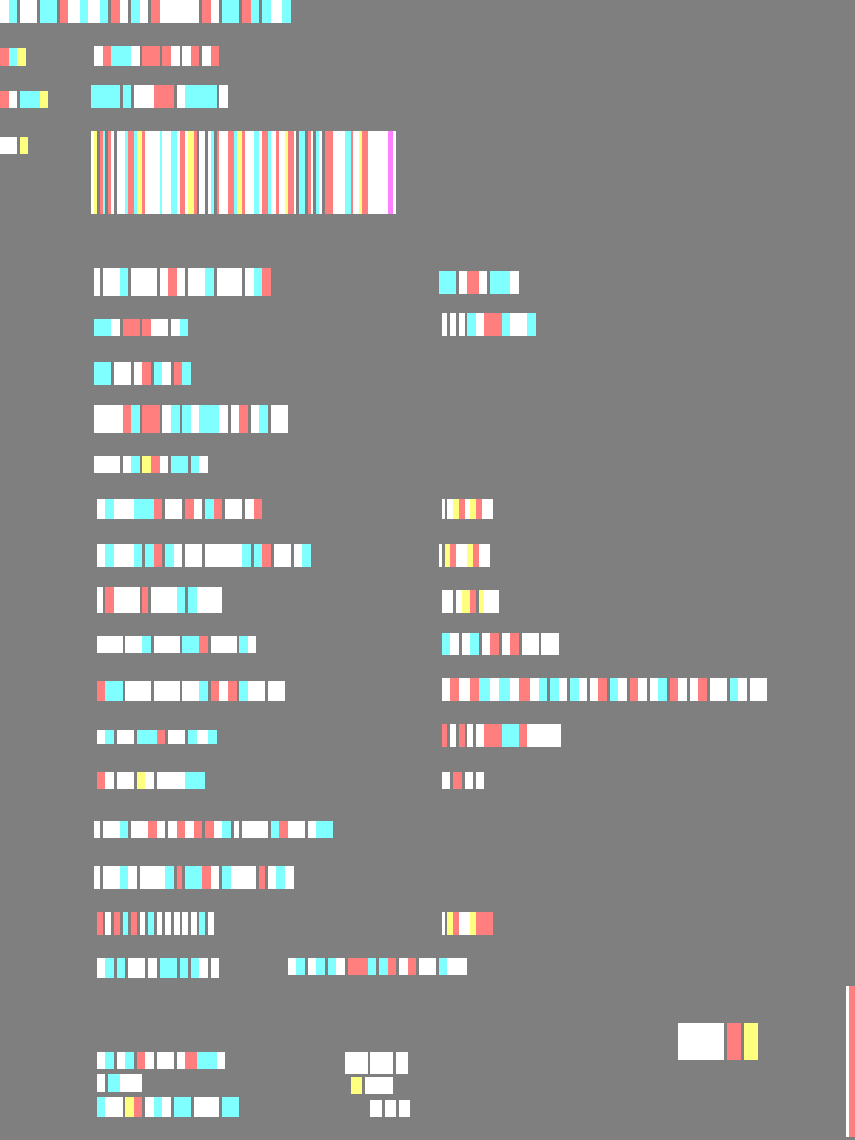}
\end{tabular}
&\begin{tabular}{c}\includegraphics[width=2cm, height=3cm]{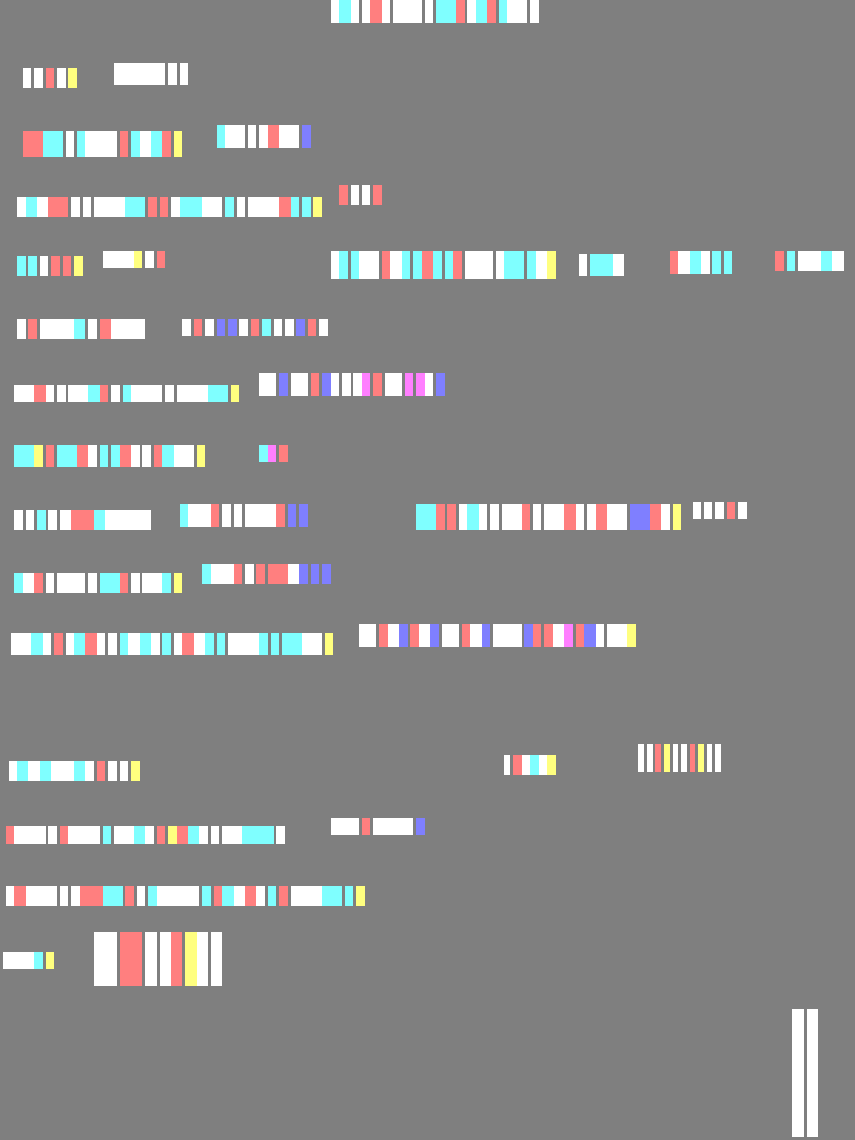}
\end{tabular}\\

Heatmap
&\begin{tabular}{c}\includegraphics[width=2cm, height=3cm]{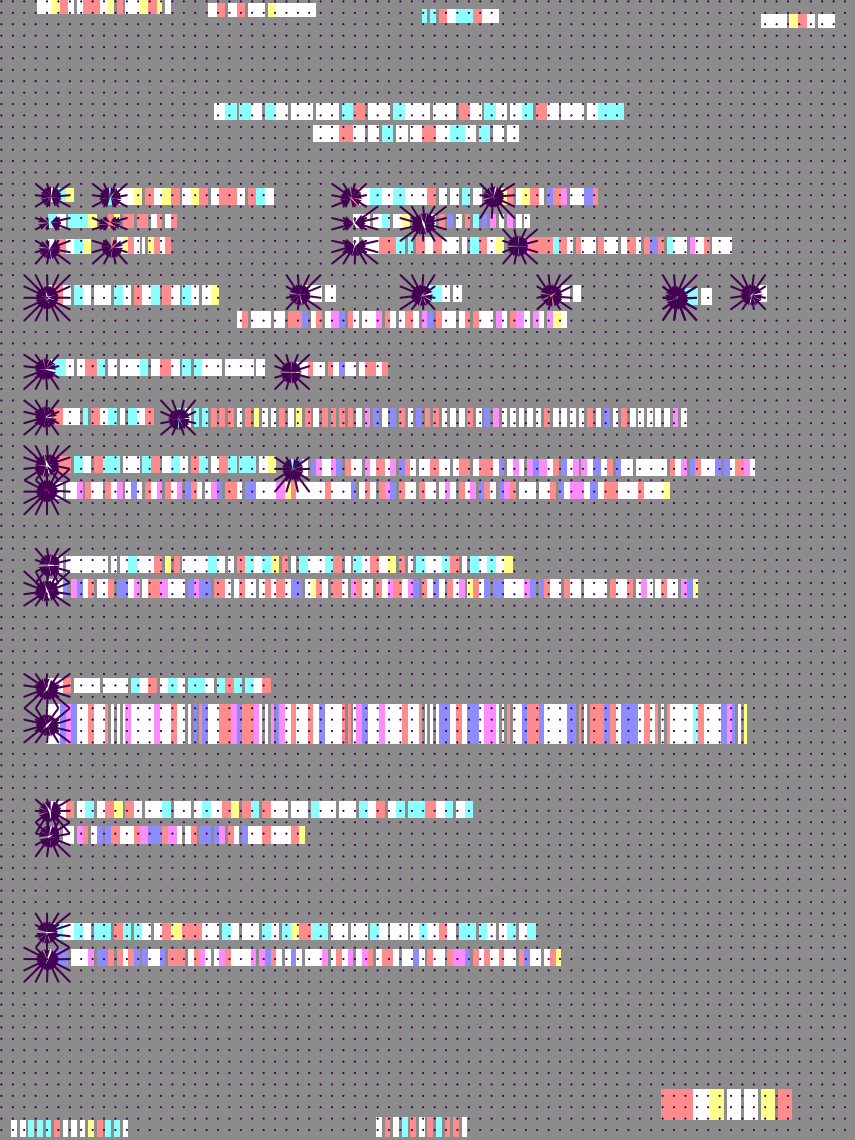}
\end{tabular}
&\begin{tabular}{c}\includegraphics[width=2cm, height=3cm]{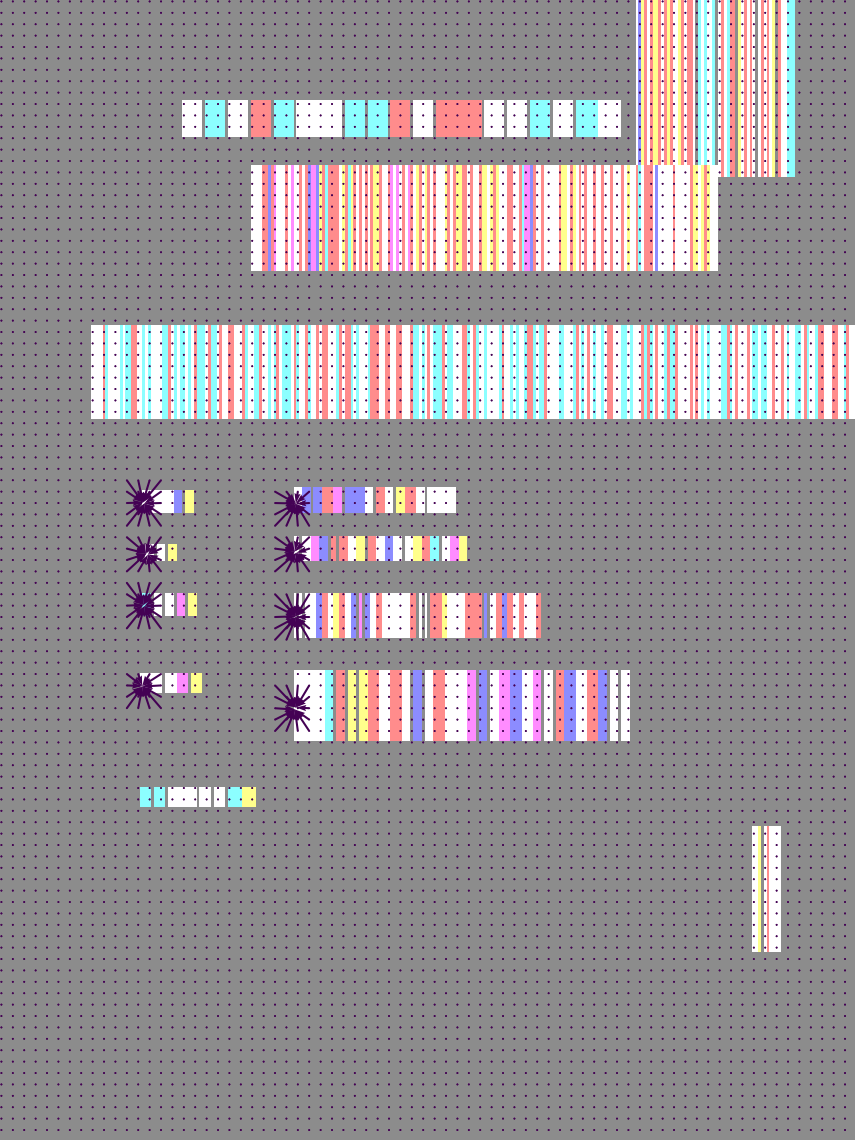}\end{tabular}
&\begin{tabular}{c}\includegraphics[width=2cm, height=3cm]{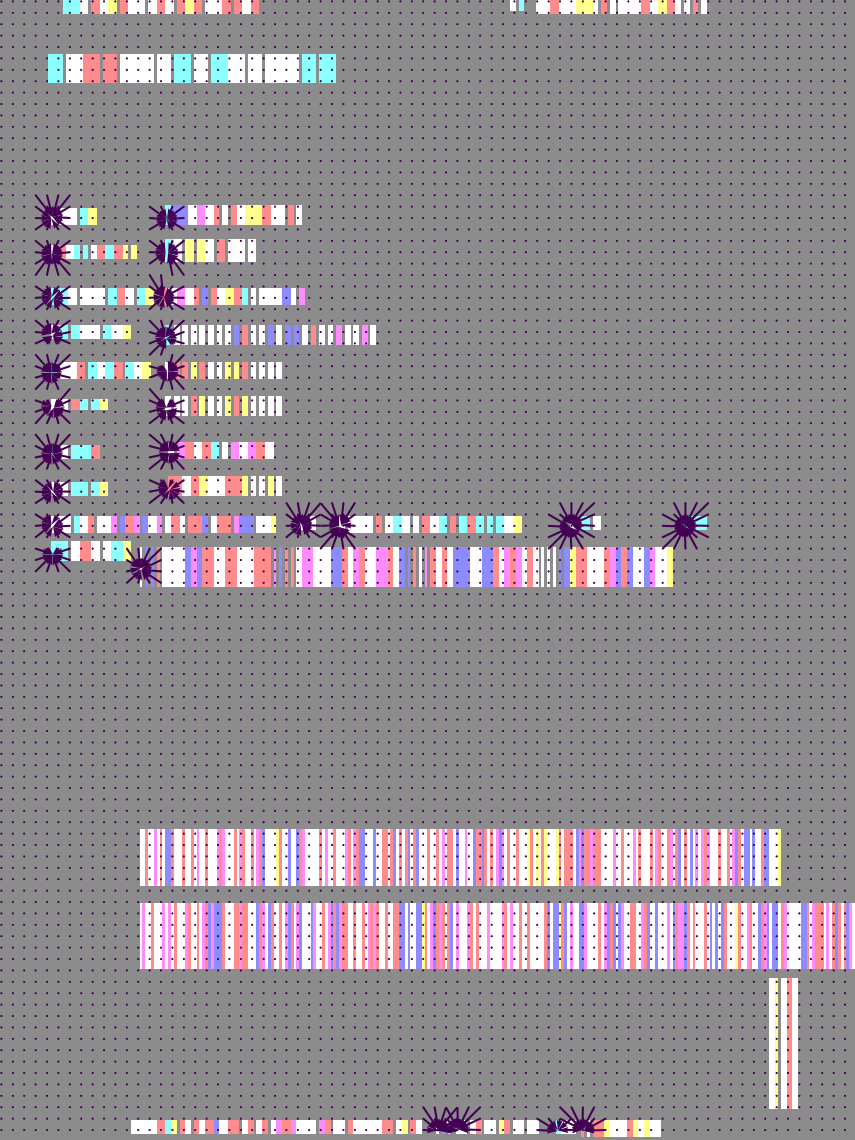}\end{tabular}
&\begin{tabular}{c}\includegraphics[width=2cm, height=3cm]{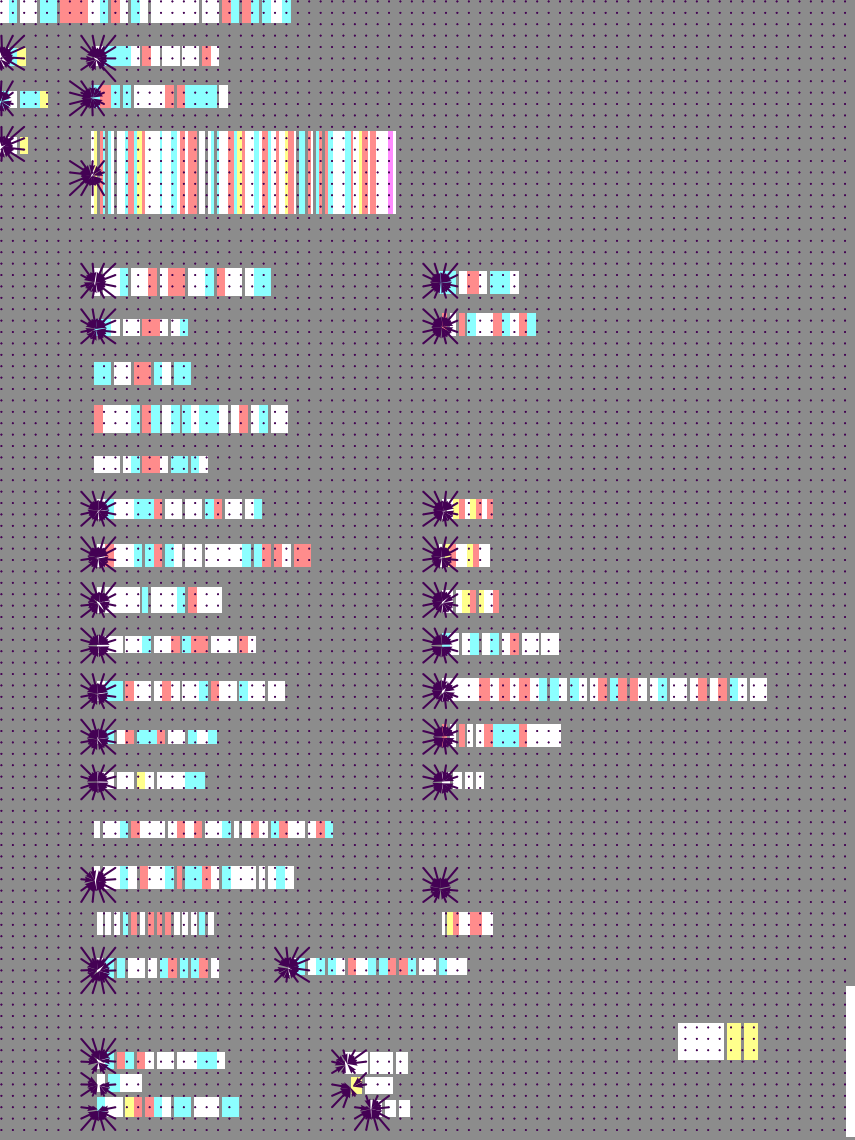}\end{tabular}
&\begin{tabular}{c}\includegraphics[width=2cm, height=3cm]{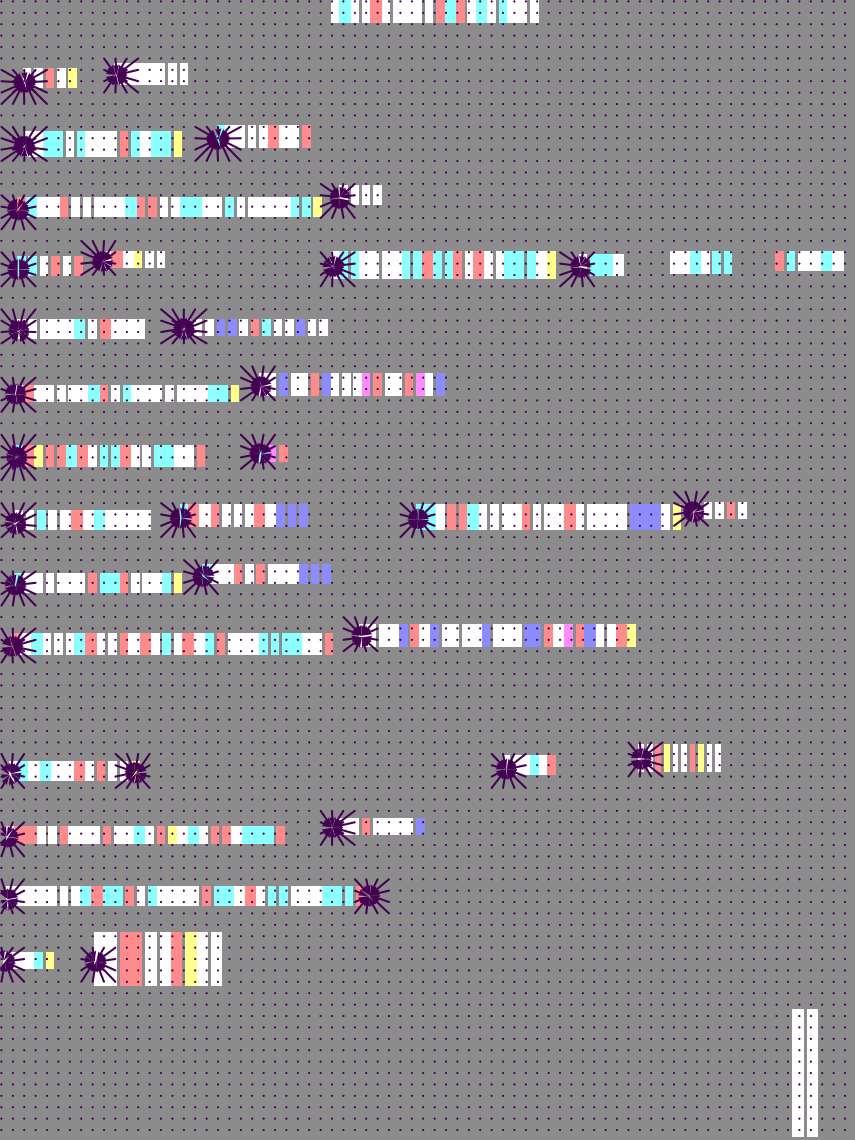}
\end{tabular}\\

Output                     
&\begin{tabular}{c}\includegraphics[width=2cm, height=3cm]{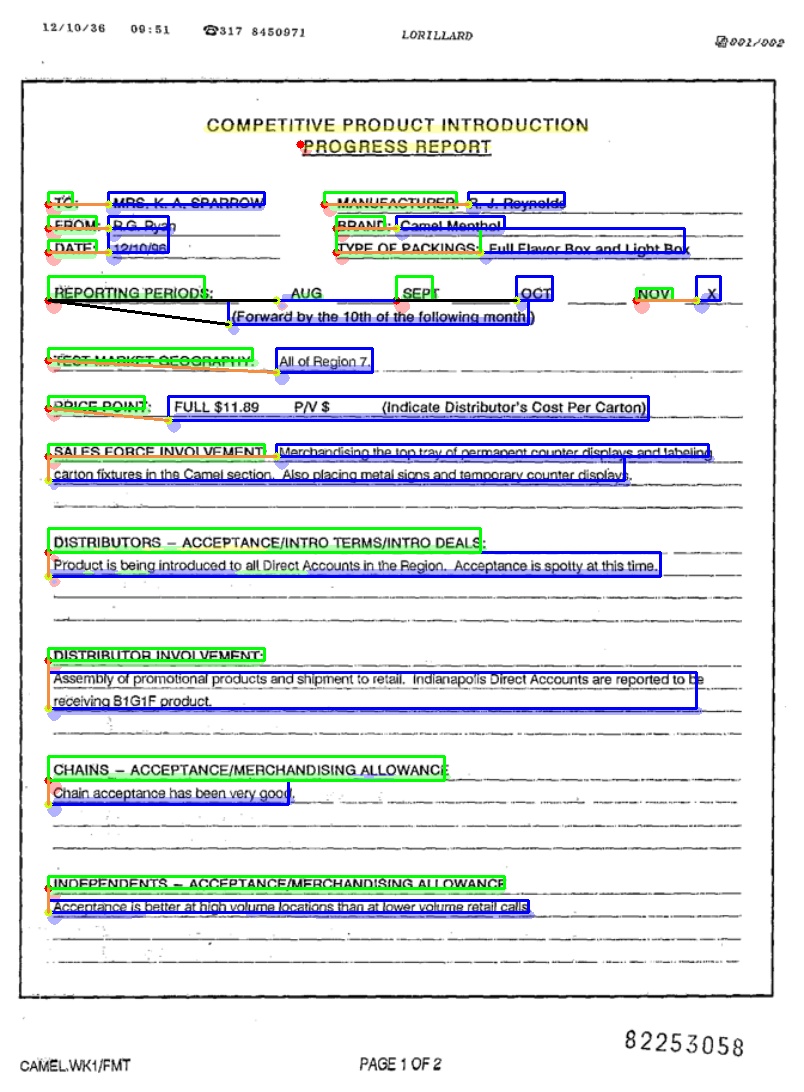}
\end{tabular}
&\begin{tabular}{c}\includegraphics[width=2cm, height=3cm]{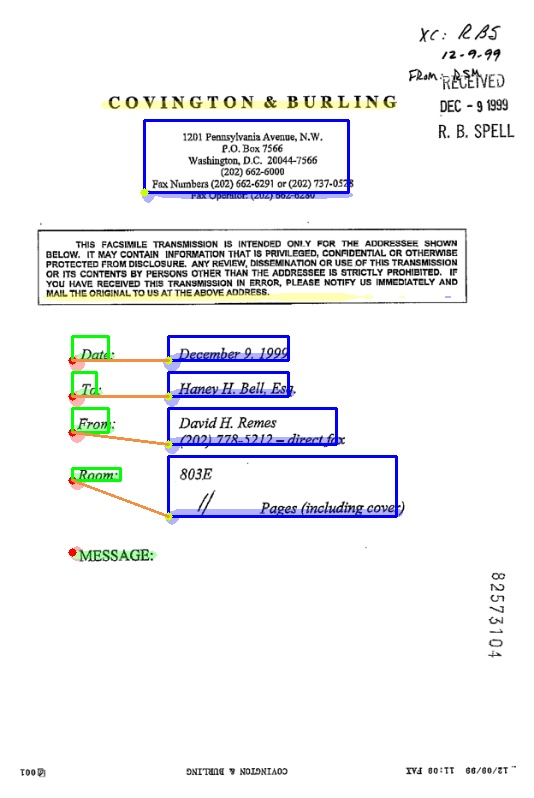}
\end{tabular}
&\begin{tabular}{c}\includegraphics[width=2cm, height=3cm]{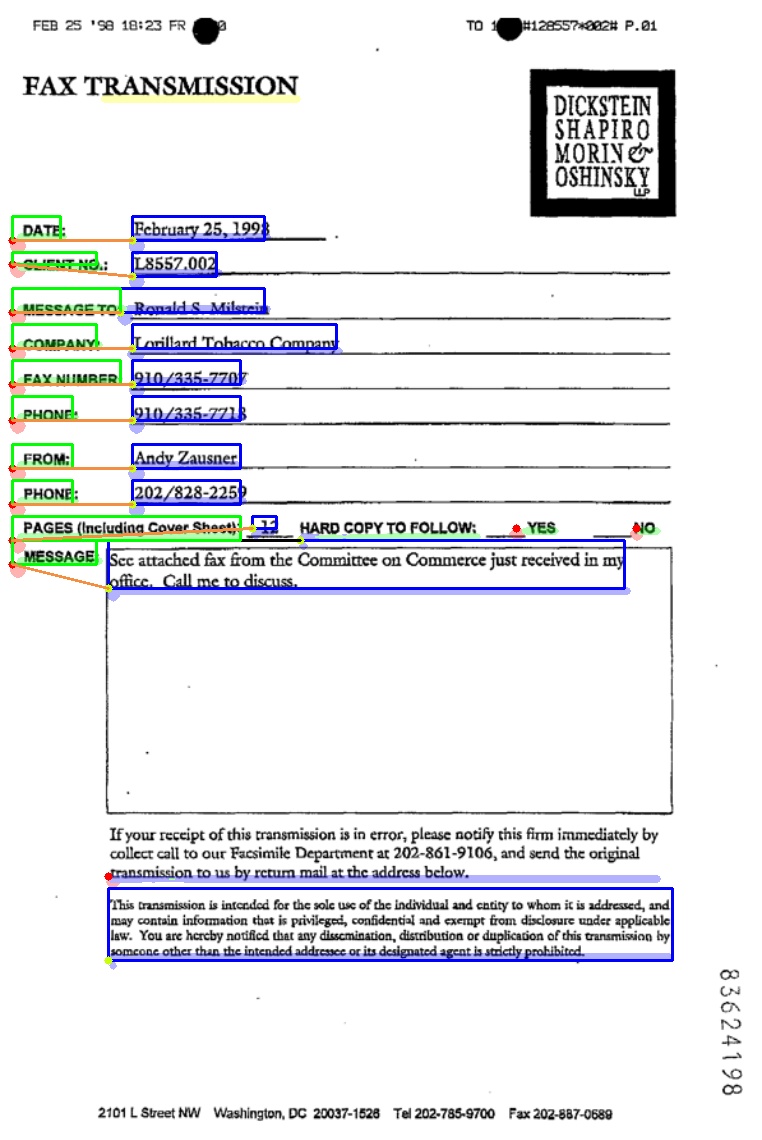}
\end{tabular}
&\begin{tabular}{c}\includegraphics[width=2cm, height=3cm]{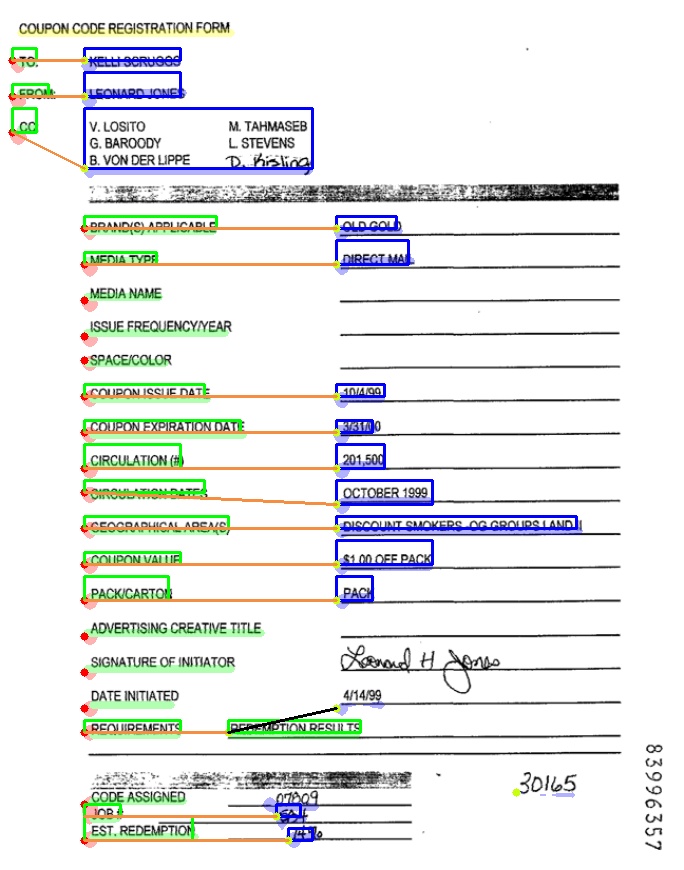}
\end{tabular}
&\begin{tabular}{c}\includegraphics[width=2cm, height=3cm]{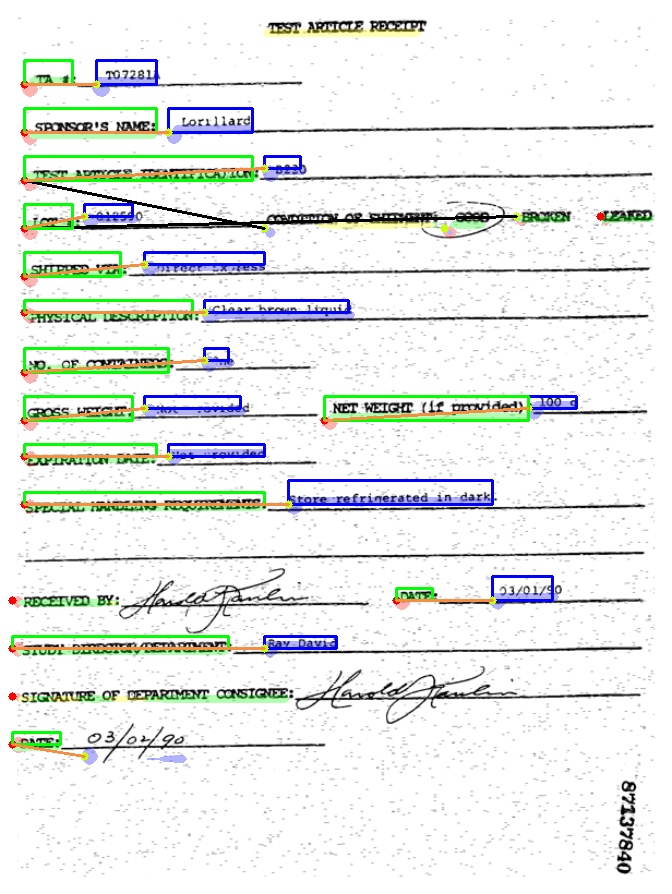}
\end{tabular}
\end{tabular}
\end{table*}

We visualize the final output by images in table \ref{table:resultofmodel}. The first and second row are input and the char-grid of images after pre-processing respectively. The result of model can be visualized in the $Output$ row. As we can see, the $Heatmap$ row shows the entity locations through bottom-left points, while the output row shows the full results of our model. The header, question, and answer are drawn by yellow, green, and blue segmentation masks respectively. The linking between entities (if have) are visualized through orange lines.
These lines start from the bottom-left points of entities labeled as the answer to the bottom-left points of entities called the question. Our model has this ability because from keypoint detection of each entity, PAFs are likely to link these text-regions by using a directed vector.

In order to benchmark the performance of our model, we evaluate it with other methods on FUNSD dataset \cite{Jaume2019}. All results are measured by F1-score.

\textbf{Entity Labeling Task}
Table~\ref{table:resultofclassification} shows the comparison between MSAU-PAF and the baselines. 
MSAU-PAF model achieves the highest F1-score 0.83, which significantly improves the accuracy compared to MLP \cite{Jaume2019} the baseline model - MSAU \cite{Madashi2019} by 0.26 and 0.1 respectively. On top of that, our model outperforms the state of the art LayoutLM \cite{Xu2019} by a large margin.
The improvement in accuracy hints that the supervision from Entity Linking can be one of the factors improving the accuracy of the model in comparison with segmentation-only supervision of MSAU \cite{Madashi2019}.
It is noted that the MLP baseline \cite{Jaume2019} utilized a much more sophisticated pre-trained model than MSAU and MSAU-PAF (BERT \cite{Devlin2018} is proven to be state-of-the-arts in many tasks from the aforementioned paper in comparison with one-hot character encoding). 
The improvement in MSAU and MSAU-PAF accuracy is potential because of the model making better use of spatial information and char-gird embeddings, as mentioned in \cite{Madashi2019}. Meanwhile, the LayoutLM mainly relies on the image embeddings via utilizing bounding box regions in the prior step, which appears to fail to capture well the spatial and relational information among entities.

\begin{table}[ht]
\centering
\caption{Result of Entity Labeling: The comparison of baseline models and MSAU-PAF on Entity Labeling task of FUNSD dataset, measured by F1-score. 
\label{table:resultofclassification}}
\begin{tabular}{|l|c|}
\hline
\multicolumn{1}{|c|}{Methods} & \multicolumn{1}{l|}{Metric (F1 score)} \\ \hline
MLP (BERT embedding) \cite{Jaume2019}          & 0.57                                   \\ \hline
MSAU (one-hot character) \cite{Madashi2019}      & 0.73                                   \\ \hline
$LayoutLM_{BASE}$ \cite{Xu2019}                  & 0.79                                    \\ \hline
MSAU-PAF (one-hot character)  & \textbf{0.83}                                   \\ \hline
\end{tabular}
\end{table}

\textbf{Entity Linking Task}
With this task, MSAU-PAF continues outperforming other methods (see in Table~\ref{table:resultoflinking}). 
It is observable that the F1-score of GraphCNN and MLP baselines are low, this can be due to the effect of highly dense and imbalanced class distribution in conjunction with Deep-learning based matching. The low F1 scores come mainly from low precision (model predicts most of non-link for most of the pairs).
The same hypothesis has been made in the original paper \cite{Jaume2019}.
In contrast, the PIF-PAF heads could particularly handle well in densely and occluded documents via effectively encoding fine-grained information and confidently predicting confidence association between two entities with composite field structure. 
This leads to 0.71 and 0.62 improvement in F1-score from MLP and GraphCNN with Link prediction respectively. It is noted that other baselines are assumed to already know the ground-truth of the class of entities, while our model only uses the text-line regions and optical character.  

Compared to the Heuristic method with Ground-truth entities class, the MSAU-PAF seems to have a little lower F1-score. This can be because of the use of the ground-truth label for each entity, while in the MSAU-PAF experiment, we combine both two tasks and the result is the joint probability of Entity Labeling and Entity Linking. The hypothesis is verified when we applied MSAU-PAF middle output as input to the heuristic rules.

\begin{table}[ht]
\centering
\caption{Result of Entity Linking: The comparison of baseline models and MSAU-PAF on Entity Linking task of FUNSD dataset, measured by F1-score.  
\label{table:resultoflinking}}
\begin{tabular}{|l|c|}
\hline
\multicolumn{1}{|c|}{Methods} & \multicolumn{1}{l|}{Metric (F1 score)} \\ \hline
MLP (BERT embedding + Positional feat) \cite{Jaume2019}     & 0.04 \\
(Ground-truth entities class) & \\
\hline
GraphCNN with Link prediction            & 0.13              \\ 
(Ground-truth entities class) & \\
\hline
Heuristics (with MSAU-PAF middle output) & 0.64              \\ \hline
Heuristics (with Ground-truth entities class) & 0.80                \\ \hline
MSAU-PAF (Joint classification and link prediction)   & \textbf{0.75}              \\ \hline
\end{tabular}
\end{table}

\begin{table}[ht]
\centering
\caption{Ablation study on CoordConv and Corner Pooling  
\label{table:ablation}}
\begin{tabular}{|l|c|c|}
\hline
Model   & Entity Labeling & Entity Linking \\ & (F1-Score) & (F1-Score) \\ \hline
MSAU-PAF                              & 0.80                                                                    & 0.72                                                                 \\ \hline
MSAU-PAF + CoordConv                  & 0.82                                                                   & 0.74                                                                      \\ \hline
MSAU-PAF + Corner Pooling             & 0.81                                                                 & 0.73                                                                     \\ \hline
MSAU-PAF + CoordConv + Corner & \textbf{0.83} & \textbf{0.75} \\
Pooling & & \\ \hline
\end{tabular}
\end{table}
\subsection{Ablation study}

The MSAU-PAF is also leveraged by two other components: Coordinate Convolution (CoordConv), and Corner pooling. 
To verify the effectiveness of these modules, the experiment is set up by testing MSAU-PAF as the baseline, following the hyperparameter settings in Subsection \ref{subsec:training_detail}.
After that, each of the components is added and retested in both tasks.
The results of these modules applied in MSAU-PAF architecture is shown in Table \ref{table:ablation}.
It can be observed that these component has improved model performance over vanilla MSAU-PAF.

\textbf{Ablation study for CoordConv}
The results of the experiments hint at the effectiveness of adding translational information by using CoordConv \cite{Liu2018a}. As shown in Table \ref{table:ablation}, when making the convolution access to its own input coordinates through concatenating extra coordinate channels, MSAU-PAF with Coordinate Convolution experienced a total gain of 0.02 in F1 score in both Entity Labeling and Entity Linking. 
It should be emphasized that using two or more CoordConvs does not bring a noticeable improvement in the experiment (less than 0.01 improvement in F1 on both tasks), so it is omitted in Table \ref{table:ablation}.
Thus, it is suggested that a single CoordConv layer will ensure both decent performance while avoiding additional computational cost.

\textbf{Ablation study for Corner Pooling}
Corner Pooling is another additional component integrated, situating between MSAU and PIF-PAF.
It is expected from the aforementioned paper \cite{Law2018} that Corner Pooling will help aggregate information better in long distance for key points detection. 
This has also been hinted by the experiment result by an increase of 0.01 in F1 score in Entity Labeling task and 0.01 in Entity Linking.
It is noted that combining CoordConv and Corner Pooling can even increase the model performance further: The last row shows the results that we test on combining all improvements, which improves the F1 score by 0.03 by the total in both Entity Labeling and Entity Linking.

\section{Conclusion}
We have proposed the MSAU-PAF - an end-to-end deep architecture for form understanding tasks. Our model extends the original MSAU architecture with link-prediction between entities by integrating the PIF-PAF mechanism and other enhancements for exploiting spatial correlation. Experiment results on the FUNSD dataset show significant improvement in comparison with other baselines. In the future, we wish to extend MSAU-PAF to other document structure parsing problems, i.e table structure recognition, document de-warping.





%

\bibliographystyle{unsrt}  
\bibliography{library}
\end{document}